\documentclass{article}

\usepackage{microtype}
\usepackage{graphicx}
\usepackage{subfig}
\usepackage{booktabs} % for professional tables
\usepackage{multirow}
\usepackage{xspace}
\usepackage{hyperref}

\newcommand{\ours}{DET\xspace}
\newtheorem{hypothesis}{Hypothesis}

\usepackage[accepted]{icml2022}
\usepackage{amsmath,amsfonts,bm}

\def\eqref#1{equation~\ref{#1}}
\def\1{\bm{1}}

\def\rvb{{\mathbf{b}}}
\def\rvc{{\mathbf{c}}}
\def\rvd{{\mathbf{d}}}
\def\rve{{\mathbf{e}}}

\def\rvh{{\mathbf{h}}}

\def\rvr{{\mathbf{r}}}

\def\rvv{{\mathbf{v}}}

\def\mA{{\bm{A}}}
\def\mB{{\bm{B}}}

\def\mH{{\bm{H}}}

\def\mK{{\bm{K}}}

\def\mQ{{\bm{Q}}}

\def\mV{{\bm{V}}}
\def\mW{{\bm{W}}}
\def\mX{{\bm{X}}}
\def\mY{{\bm{Y}}}

\DeclareMathAlphabet{\mathsfit}{\encodingdefault}{\sfdefault}{m}{sl}
\SetMathAlphabet{\mathsfit}{bold}{\encodingdefault}{\sfdefault}{bx}{n}

\def\gE{{\mathcal{E}}}

\def\gG{{\mathcal{G}}}

\def\gM{{\mathcal{M}}}
\def\gN{{\mathcal{N}}}

\def\gV{{\mathcal{V}}}

\def\sR{{\mathbb{R}}}

\newcommand{\sigmoid}{\sigma}

\icmltitlerunning{Unleashing the Power of Transformer for Graphs}

\begin{document}

\twocolumn[
\icmltitle{Unleashing the Power of Transformer for Graphs}

\icmlsetsymbol{equal}{*}

\begin{icmlauthorlist}
\icmlauthor{Lingbing Guo}{zju}
\icmlauthor{Qiang Zhang}{zju}
\icmlauthor{Huajun Chen}{zju}
\end{icmlauthorlist}

\icmlaffiliation{zju}{College of Computer Science and Technology, Zhejiang University, Hangzhou, China}

\icmlcorrespondingauthor{Qiang Zhang}{qiang.zhang.cs@zju.edu.cn}

% You may provide any keywords that you
% find helpful for describing your paper; these are used to populate
% the "keywords" metadata in the PDF but will not be shown in the document
\icmlkeywords{Machine Learning, ICML}

\vskip 0.3in
]

\printAffiliationsAndNotice{} % otherwise use the standard text.

\begin{abstract}
Despite recent successes in natural language processing and computer vision, Transformer suffers from the scalability problem when dealing with graphs. The computational complexity is unacceptable for large-scale graphs, e.g., knowledge graphs. One solution is to consider only the near neighbors, which, however, will lose the key merit of Transformer to attend to the elements at any distance. In this paper, we propose a new Transformer architecture, named dual-encoding Transformer (DET). DET has a structural encoder to aggregate information from connected neighbors and a semantic encoder to focus on semantically useful distant nodes. In comparison with resorting to multi-hop neighbors, DET seeks the desired distant neighbors via self-supervised training. We further find these two encoders can be incorporated to boost each others' performance. Our experiments demonstrate DET has achieved superior performance compared to the respective state-of-the-art methods in dealing with molecules, networks and knowledge graphs with various sizes.
\end{abstract}

\section{Introduction}
\label{sec:intro}

Transformer has become one of the most prevalent neural models for natural language processing (NLP) \cite{Transformer,Bert}. The self-attention mechanism leveraged by Transformer has already been extended to graph neural network (GNN) models, e.g., GAT~\cite{GAT} and its variants~\cite{RDGCN,CompGCN,SuperGAT,HittER}. Nevertheless, these models only consider the 1-hop neighbors, which violates the original intention of Transformer that attends to the elements at distant positions. In this regard, they are graph convolutional networks (GCNs) with learnable attention weights for aggregating 1-hop neighborhood.

Recently, Graphormer \cite{Graphormer} starts to leverage the standard Transformer architecture for graph representation learning and has achieved state-of-the-art performance on many benchmarks. Different from large-scale knowledge graphs (KGs) or networks, the input graph for the graph property prediction tasks are limited with small sizes (e.g., small molecules). Graphormer is thus inapplicable on large-scale graphs. 
The same problem also appears in the computer vision (CV) area, yet has recently been tackled by patching pixels to patches and then to windows in a hierarchical fashion~\cite{ViT,SwinTransformer}. These works inspire us to explore the possibility of using one universal Transformer architecture as the general backbone to model different sizes of graphs.

\begin{figure*}[t]
	\centering
	\includegraphics[width=.96\linewidth]{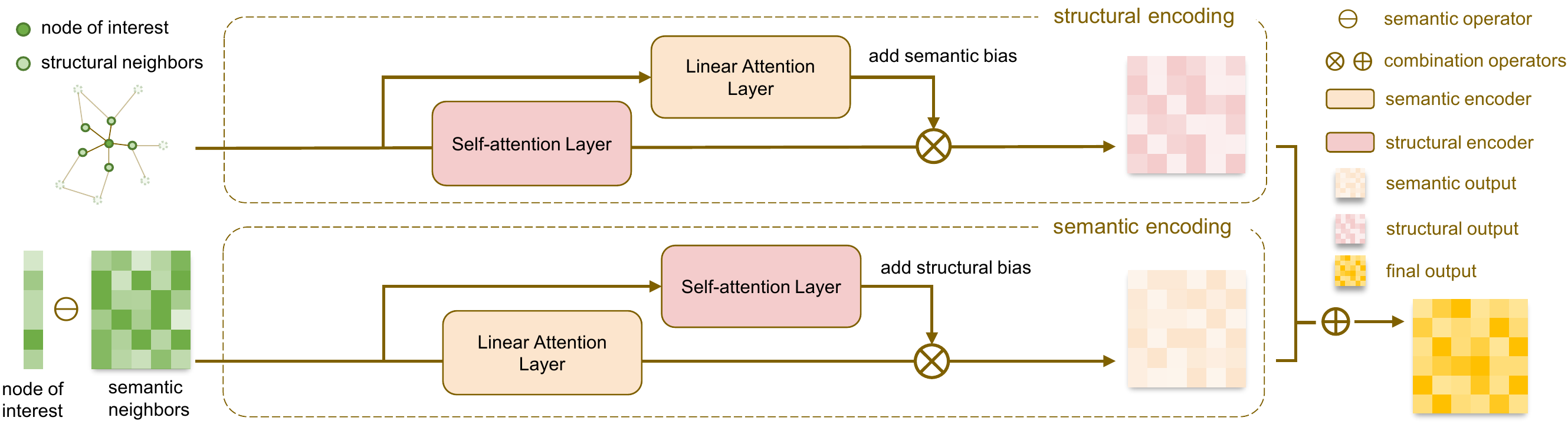}
	\caption{Overview of \ours. Structural neighbors are nodes connected with the node of interest on the graph, while semantic neighbors are nodes with similar semantics to the node of interest. The two encoders focus on encoding different aspects of neighboring information, and thus are capable of complementing each other.}
	\label{fig:arch}

\end{figure*}

How to attend to the nodes at arbitrary distance? Is it possible to directly aggregate the embeddings of multi-hop neighbors? Yes, but it is not necessary. As far as we know, there are few GCN-based or GAT-based methods considering the three or more-hop neighbors. Even the improvement brought by the two-hop neighbors is very limited \cite{AliNet,HittER}. Using one-hop neighbors is sufficient to achieve promising performance in most cases.

However, are all multi-hop neighbors useless? Definitely not, we just lack an effective approach to seek the useful distant neighbors from the enormous noise. In this paper, we propose dual-encoding Transformer (\ours) to tackle the above problem. In \ours, we consider two types of neighbors, i.e., structural neighbors and semantic neighbors. Structural neighbors are usually the one-hop neighbors leveraged by existing GNN models~\cite{CompGCN,SuperGAT,HittER}. Semantic neighbors, on the other hand, are the semantically close neighbors that are not directly connected to the node of interest. 

Figure~\ref{fig:arch} shows the basic idea of \ours to let two encoders continuously improve each other. For structural encoding, we use the standard self-attention layer to encode the graph structure information. The attention scores of semantic encoder is incorporated as the semantic attention bias to assist the structural encoder. We then model semantic encoding as the dual process, with the semantic encoder as the main attention model and the structural encoder as an assistant model. Therefore, \ours ensures efficient neighborhood aggregation while encouraging global node connections. 
%We then combine the outputs of two encoders to obtain the final embedding for the downstream tasks.

The key is learning a semantic encoder to identify useful distant nodes as semantic neighbors. The one-hop neighbors usually contribute the most valuable information to represent the node of interest. Therefore, the problem can be converted to find the distant nodes that are as important as the one-hop neighbors. Towards this end, we add a contrastive loss in structural encoding. We use the near neighbors as positive examples and randomly sampled distant nodes as negative examples, such that the training process can be fully unsupervised. 

Another key design is the semantic operator $\ominus$ used to estimate the semantic score between the node of interest and others. We consider a learnable similarity function rather than directly feeding the encoder with raw embeddings. This is because that the similarity is one of the most important metrics for clustering embeddings, which compels the encoder to value the semantically close yet structurally distant nodes. The idea of considering semantically close neighbors is similar to MSA Transformer~\cite{MSATransformer} and AlphaFold 2~\cite{AlphaFold2} for proteins, except that the family members (i.e., semantic neighbors in this paper) are obtained from self-supervised learning rather than querying the genetic datasets. 

The proposed \ours is capable of achieving superior performance on various learning tasks oriented to graphs with different sizes: (1) For the graph property prediction task, \ours outperforms the best-performing methods on the PCQM4M-LSC~\cite{LSC} and ZINC~\cite{ZINC} datasets; (2) For the node classification task, \ours has competitive or better performance than the state-of-the-art attention-based methods, on the Cora, CiteSeer, PubMed and PPI benchmarks~\cite{Cora,PPI}; (3) For the KG completion (a.k.a., entity prediction) task, \ours achieves the state-of-the-art performance on both FB15K-237~\cite{Node+LinkFeat} and WN18RR~\cite{ConvE} datasets. We also present in-depth model analysis to provide more insights. 

\section{Related Works}
\label{sec:related_work}

We split the related literature into two parts: self-attention and position embeddings.

\subsection{Self-attention}
Self-attention-based neural models, such as Transformer, have recently become the \emph{de facto} choice in NLP tasks, ranging from language modeling and machine translation~\cite{Transformer,Bert} to question answering~\cite{BertQA} and sentiment analysis~\cite{BertSA}. Transformer has significant advantages over conventional sequential models like recurrent neural networks (RNNs)~\cite{RNN,LSTM} in both scalability and efficiency. 

When modelling graph data, another well-used self-attention mechanism is proposed by GAT~\cite{GAT}, which we call the linear attention in this paper. Compared with the dot-product calculation in the original implementation, the linear attention performs concatenation of query and key nodes before a linear transformation. Furthermore, it only considers the one-hop neighbors for the query node, which is efficient when ranking candidate nodes on large graphs. Therefore, it has become the primal choice for modeling graph data~\cite{RDGCN,AliNet,CompGCN,CGAT,SuperGAT}. 

In this paper, we use the dot-product attention to encode the structural neighbors and the modified linear attention to encode semantic neighbors.

\subsection{Position Embedding}

Position embeddings are one of the most important attributes to Transformer. Transformer variants adapted to different areas usually customize this attribute. For example, ViT~\cite{ViT} sequentially indexes the patches and encodes the indices as 1D position embeddings. SwinTransformer~\cite{SwinTransformer} proposes the 2D-aware relative position biases, which employs a learnable matrix to record pairwise patch position information in the window. 

In addition to the position information, other prior knowledge can also be injected as attention biases or embeddings into Transformer. For example, Graphformer~\cite{Graphormer} encodes centrality and shortest path distance into embeddings, and then incorporates them as ``position embeddings'' into Transformer. HittER~\cite{HittER} adds the edge type  (i.e., relation) information of KGs when encoding entity embeddings. 

Specially, MSA Transformer~\cite{MSATransformer,AlphaFold2} leverages the multiple sequence alignment (MSA) information, which is closely related to our work. When aggregating the information of one site, it does not only consider the others along the protein sequence, but also the same site within the evolutionary family members. The semantic encoding proposed in this paper also exploits such ``family'' information, except that the family members is selected based on the semantic scores.

\section{Methodology}
\label{sec:methodology}
In this section, we will illustrate the detail and insight of \ours. We start from the preliminaries, and then introduce the dual-encoding process leveraged by \ours, after which we illustrate how to train the semantic encoder to obtain the semantic neighbors. Finally, we present the implementation of the whole architecture.
\subsection{Preliminaries}
We first introduce the terminologies and notations that will be used in the following sections.

\paragraph{Graph} We define a graph as $\gG=(\gV,\gE)$, where $\gV=\{v_1, v_2, ..., v_n\}$ is the node set, and $\gE=\{e_1, e_2, ..., e_m\}$ is the edge set. $n$ and $m$ denote the numbers of nodes and edges, respectively. In practice, different tasks often have more complicated graph structures. For example, molecular graphs and KGs have typed edges (i.e., chemical bonds and relations). We do not discuss these details in the main paper, and follow the general setting to process these specific features~\cite{Graphormer,HittER}. 

\paragraph{GNN and Self-attention} Without loss of generality, we define a GNN as a neural network that learns a group of weights to aggregate the embeddings of the one-hop or multi-hop neighbors for the node of interest. In this sense, self-attention can be naturally treated as a GNN model. Let $\mQ \in \sR^{n\times h}$, $\mK \in \sR^{n\times h}$, $\mV \in \sR^{n\times h}$ denote the query, key, and value matrices, respectively. $h$ denotes the hidden-size. In this paper, they are 
the same node embedding matrix. Self-attention calculates the attention scores as follows:
\begin{equation}
    \mA = \textit{Softmax}(\frac{\mQ \mK^\top}{\sqrt{h}}),
\end{equation}
where $\mA\in \sR^{n\times n}$ records the node-to-node attention scores. We then aggregate the node embeddings with the following equation:
\begin{equation}
    \mH = \mA\mV,
\end{equation}
where $\mH \in \sR^{n\times h}$ is the output embedding matrix, with each row denoting the output embedding of a node in $\gV$.

\paragraph{Linear Attention} The computational complexity of the above dot-product implementation is $\omega(n^2)$ (without considering hidden-size). As the number of nodes increases, the cost becomes unacceptable. GAT~\cite{GAT} proposes a linear self-attention implementation to mitigate this problem by only considering the one-hop neighbors:
\begin{align}
	\label{eq:att_w}
	\mB_{ij} = \sigma(\rvb^\top ( \rvv_i \mW \,\rVert\, \rvv_j \mW)),
\end{align}
where $\mB_{ij}$ denotes the attention score from the node of interest $v_i$ to a neighbor $v_j$. $\rvv_i$, $\rvv_j \in \sR^{h}$ are the embeddings of $v_i$ and $v_j$, respectively. $\rvb \in \sR^{h}$ and $\mW \in \sR^{h\times h}$ are weight vector and matrix, respectively. $\sigma$ is the activation function and $\rVert$ denotes the concatenation operation. The linear attention does not consider the correlations within neighbors, and thus its computational complexity is cut down from $\Omega(n^2)$ to $\Omega(n+m)$. 

\subsection{Dual Encoding}
The well-used two attention mechanisms have different pros and cons. To fulfill our goal of designing a universal graph Transformer architecture, we propose to combine their advantages to effectively and efficiently encode graphs with different sizes.

\paragraph{Structural Encoding}
The standard dot-product attention can be easily extended on the small graphs. We add a virtual node $v_c$ \cite{Bert} as the context node connected with all nodes in $\gG$. Then, the output representation for $v_c$ can be regarded as the representation of $\gG$. For the large graphs like KGs or networks, we can perform self-attention on the one-hop ego graph $\gG_i$ given the node of interest $v_i$. Therefore, the output embedding for node $v_i$ is
\begin{align}
    \label{eq:structual}
    \rvh^{\text{st}}_i &= \sum_{v_j\in \{v_i\} \cup \gN(v_i)} \mA_{cj} \rvv_j,
\end{align}
where $\rvh^{\text{st}}_i$ denotes the output of structural encoder for $v_i$. $\mA_{cj}$ denotes the attention score for the context node $c$ to the neighbor $v_j$. $\gN(v_i)$ denotes the structural one-hop neighbor set for $v_i$. With multiple stacked layers, the correlations within the one-hop neighbors will be leveraged.

Therefore, we can use one universal encoder layer to model the graph structural information to obtain the node representation and graph representation. Note that, the encoder itself does not consider the difference of nodes in the graph structure (e.g., distance to the node of interest). Follow the exiting works~\cite{Graphormer,HittER}, we accordingly add the centrality, relation type, or shortest distance path information as special position embeddings to the encoder. The details can be found in Appendix~\ref{app:pos}.

\paragraph{Semantic Encoding}
The structural features may be less discriminative to identify a neighbor. For example, if the node of interest has a large amount of neighbors, it is inevitable that many neighbors will have similar or identical structural features (e.g., shortest path distance to $v_i$). This problem is even more serious when we only consider the one or two hop neighbors. However, if we consider more than two hop neighbors, the sheer quantity of available information will overwhelm the neural network.

In this paper, we seek the distant neighbors not based on paths but the semantic scores estimated by the semantic encoder. Specifically, the semantic similarity is estimated by a learnable neural function $f_s: \sR^h \times \sR^h \rightarrow \sR$:
\begin{align}
    \label{eq:semantic_encoder}
    f_s(\rvv_i, \rvv_j) &= \sigmoid(\rvv_i \ominus \rvv_j) \nonumber\\ 
                        &= \sigmoid((\mW_s(\rvv_i - \rvv_j)+ b_s),
\end{align}
where $\ominus$ is the semantic difference operator. The choice of $\ominus$ is flexible, as long as it can reflect the similarity between $\rvv_i$ and $\rvv_j$. Here, we use a weighted $L1$ distance as $\ominus$, which is simple and can be easily extended to matrix operation. $\mW_s \in \sR^{1\times h}$ and $b_s \in \sR$ (for single head attention) are the weight matrix and bias, respectively.

\paragraph{Dual-encoding Transformer}
In \ours, the structural encoder encodes the graph structural information into node embeddings, while the semantic encoder calculates the attention scores based on the semantic similarity. It is worth incorporating these two encoders to achieve better performance. As shown in Figure~\ref{fig:arch}, when the structural encoder fails to estimate which neighbors are more important, the attention bias provided by the semantic encoder may be helpful. For example, in a citation network, the semantic similarity may be a more effective metric than structural distance when aggregating the neighboring papers for classification. The labels of the semantically close followers/followees are the key components to predict the label of the node of interest. For the dual process -- semantic encoding, we also expect that the structural encoder can assist the semantic encoder. For example, the structural information like centrality for a distant node is useful in encoding semantic neighbors. The dot-product attention also provides another view for aggregating semantic neighbors.

\subsection{Exploiting the Semantic Neighbors}

\begin{table}[t]
\centering
\small
\caption{Average node appearing time on FB15K-237 and WN18RR datasets, in term of hops.}
%\vskip 0.15in
\label{tab:avg_deg}
\begin{tabular}{crr}
    \toprule
    Hops                & FB15K-237  & WN18RR \\ \midrule
    1-hop & 20.3 & 2.7 \\
    2-hop & 1,781.4 & 8.9\\
    3-hop & 64,774.9& 30.5\\
    5-hop & -& 483.8\\
    \bottomrule
\end{tabular}
\end{table}

To establish a stable dual-encoding process, the key is seeking reliable semantic neighbors. In this paper, we propose a self-supervised training method to exploit the semantic neighbors of the node of interest, which is based on the following two hypotheses:
\begin{hypothesis}
The one-hop neighbors are the most informative sources to identify and represent the node of interest.
\end{hypothesis}
Intuitively, if a node appears as a neighbor of other nodes for many times, the information provided by this node may not be helpful to make the node of interest \emph{unique}. Table~\ref{tab:avg_deg} shows the average appearing times for different hops of neighbors. We can find that the two or more-hop neighbors of a node are shared by more others, which is why current GNN models rarely consider the multi-hop neighbors. This is also known as the over-smoothing problem. Therefore, it is reasonable to regard the one-hop neighbors as the most informative sources.
\begin{hypothesis}
The distant nodes with similar embeddings to that of the node of interest are important sources to the node of interest.
\end{hypothesis}
It has been proven that the information provided by the family members is useful for protein structure prediction in MSA Transformer~\cite{MSATransformer} and AlphaFold2~\cite{AlphaFold2}. Although the idea of the semantic encoder is not directly inspired by these two existing works, we actually find the close correlations between the semantic neighbors and the MSA information, which may explain why \ours works. 
\begin{figure}[t]
	\centering
	\includegraphics[width=.99\linewidth]{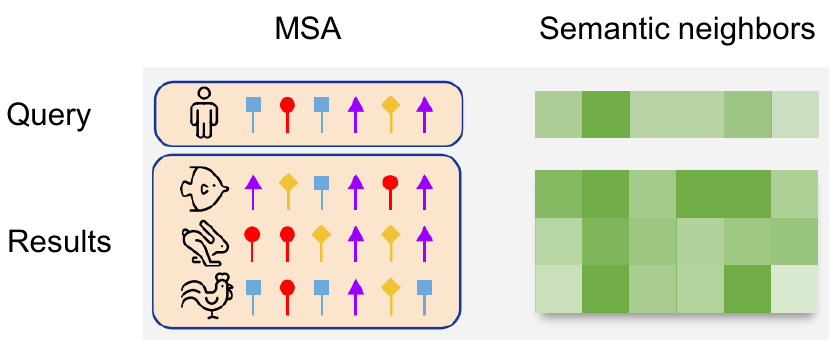}
	\caption{Comparing MSA with semantic neighbors. The left figure is sliced from~\cite{AlphaFold2}.}
	\label{fig:msa}
% 	\vspace{-1.2em}
\end{figure}

Multiple sequence alignment (MSA) can be regarded as the results of biological sequence alignment. The protein sequences in the same family are assumed to have a common ancestor or an evolutionary relationship. The semantic neighbors of the node of interest, on the other hand, are also assumed to have similar semantic characteristics to the node of interest. We compare MSA and semantic neighbors in Figure~\ref{fig:msa}. The MSA results are obtained by searching the genetic database, while the semantic neighbors are obtained by searching the node embedding set with the semantic encoder. Therefore, the semantic encoder in this sense is equivalent to a column attention layer leveraged by MSA Transformer and AlphaFold 2.

Based on the above two hypotheses, we define the semantic neighbor fetching loss as:
\begin{align}
\label{eq:sn_loss}
    \mathcal{L}_\text{sn}(v_i) = &-\frac{1}{|\gN(v_i)|}\sum_{v_j\in \gN(v_i)}{\ln(f_s(\rvv_i, \rvv_j))}\nonumber \\ 
                                     &+\frac{1}{|\gN^-(v_i)|}\sum_{v_k\in \gN^-(v_i)}{\ln(f_s(\rvv_i, \rvv_k))},
\end{align}
where $\gN^-(v_i)$ is the negative neighbor set where the negative examples are randomly sampled distant nodes to $v_i$. 
\subsection{Implementation}

\begin{algorithm}[t]
   \caption{Dual-encoding Transformer}
   \label{alg:det}
\begin{algorithmic}
   \STATE {\bfseries Input:} graph $\gG=(\gV,\gE)$, the prediction loss $\mathcal{L}_\text{main}$, structural encoder $\gM_\text{st}$ and semantic encoder $\gM_\text{se}$
   \STATE Initialize all parameters.
   \REPEAT
   \STATE Select the semantic neighbors for each node.
       \FOR{{\bfseries each} batch data $(\mX, \mY)$}
           \STATE $\mH_\text{st} \leftarrow \gM_{st}(\mX)|\gM_{se}(\mX)$
           \STATE $\mH_\text{se} \leftarrow \gM_{se}(\mX)|\gM_{st}(\mX)$
           \STATE $\mH \leftarrow H_\text{st} \oplus H_\text{se}$ 
           \STATE Compute $\mathcal{L}_\text{sn}$ according to Equation~(\ref{eq:sn_loss})
           \STATE $\mathcal{L} \leftarrow \mathcal{L}_\text{main}(\mH, \mY) +  \mathcal{L}_\text{sn}$
           \STATE Update the parameters according to $\mathcal{L}$
       \ENDFOR
   \UNTIL{the performance on validation set converges}
\end{algorithmic}
\end{algorithm}
We illustrate the implementation of \ours by Algorithm~\ref{alg:det}. We first initialize all parameters of \ours and embeddings. For better efficiency, the semantic neighbors of each node are only updated at the beginning of each epoch. Then, for each batch, the structural output embeddings are generated by the structural encoder conditioned on the semantic encoder bias, and verse visa. Finally, we combine the main prediction loss and the semantic neighbor fetching loss, and update all parameters by back-propagation.

\section{Experiment}
\begin{table}[t]
	\small
	\centering
	\caption{Graph property prediction results on PCQM4M-LSC.}
	\label{tab:main_graph_pcq}
	\begin{tabular}{lrrr}
		\toprule
		Model & \#param. & train MAE     & validate MAE \\ \midrule
		GCN & 2.0M & 0.1318   & 0.1691 \\
		DeeperGCN & 25.5M & 0.1059 & 0.1398 \\
		GraphSage & - & - & - \\
		GIN & 3.8M & 0.1203 & 0.1537 \\ 
		% GatedGCN-PE~\cite{} & - & - & - & 505,011   & 0.214\\
		% MPNN (sum)~\cite{} & - & - & -  & 480,805   & 0.145 \\
		\midrule
		GT  & 83.2M  & 0.0955 & 0.1408 \\ 
		Graphormer& 47.1M  & 0.0582 & 0.1234 \\\midrule
		\ours & 47.1M & \textbf{0.0546} & \textbf{0.1212} \\
		\bottomrule
	\end{tabular}
\end{table}

\begin{table}[t]
	\small
	\centering
	\caption{Graph property prediction results on ZINC. }
	\label{tab:main_graph_zinc}
	\begin{tabular}{lrr}
		\toprule
		Model & \#param. & test MAE \\ \midrule
		GCN & 505,079   & 0.367 \\
		GraphSage & 505,341   & 0.398 \\
		GIN & 509,549   & 0.526 \\ 
		PNA        & 387,155   & 0.142 \\
		\midrule
		GAT & 531,345 & 0.384 \\
		SAN & 508,577 & 0.139 \\
		GT  & 588,929 & 0.226 \\ 
		Graphormer & 489,321 & 0.122   \\\midrule
		\ours & 489,562 & \textbf{0.113}\\
		\bottomrule
	\end{tabular}
\end{table}

\label{sec:expr}
In this section, we conducted experiments on a variety of benchmarks, ranging from graph property prediction to node classification and KG completion, to verify the effectiveness of \ours. The source code was uploaded and will be available online. Please see Appendix~\ref{app:stat} for dataset statistics. 

\subsection{Graph Property Prediction}
\paragraph{Settings}
We evaluated \ours on the graph property prediction benchmarks PCQM4M-LSC~\cite{OGB-LSC} and ZINC~\cite{ZINC}. The former is used in recent Open Graph Benchmark Large-Scale Challenge\footnote{https://ogb.stanford.edu/kddcup2021/pcqm4m/}, while the later is a popular dataset used to evaluate graph representation learning methods.

% The training setting was mainly adopted from \cite{Graphormer}. 
We employed a 12-layer \ours with hidden-sizes $768$ and $80$, for PCQM4M-LSC and ZINC, respectively. Due to the number of nodes for each graph is very small (usually less than $50$), we directly perform attention operations on the whole graph rather than on one-hop or semantic neighbors. Therefore, we removed the semantic neighbor fetching loss in this experiment.

\paragraph{Baselines}
We compared \ours with state-of-the-art GNN methods on the dataset leader-boards: the attention-based methods SAN~\cite{SAN}, GT~\cite{GT} and Graphormer~\cite{Graphormer}; and other recently developed GNN methods GCN~\cite{GCNs}, GraphSage~\cite{graphsage}, GIN~\cite{GIN}, DeeperGCN~\cite{NeuralMessagePassing}, and PNA~\cite{PNA}.

\paragraph{Results}
Table~\ref{tab:main_graph_pcq} and Table~\ref{tab:main_graph_zinc} summarize the experimental results measured by mean average error (MAE) on two datasets. Due to the inaccessibility of the testing data on PCQM4M-LSC, we alternatively report the MAE results on training and validation sets.

Overall, \ours outperformed all the baseline methods, achieving a new state-of-the-art on both two datasets. Compared with Graphormer that only considers encoding structural information with Transformers, \ours significantly improved the performance, with 6.2\% and 7.4\% MAE decline on PCQM4M-LSC and ZINC, respectively. Furthermore, the number of model parameters was still relatively equivalent to that of baselines. We also observe that \ours had more significant advantages over other methods on ZINC. This is because the restriction of the model size on the ZINC benchmark severely limited the capability of modeling structural information. Leveraging the semantic information becomes a more efficient choice. If we do not limit the model size of \ours, it can easily achieve a MAE of $0.006$.

\subsection{Node Classification}
\paragraph{Settings}
We evaluated \ours on four benchmarks that are generally used for node representation learning. Specifically, Cora, CiteSeer, and PubMed ~\cite{Cora} are three citation network datasets used for the transductive setting, while PPI~\cite{PPI} is a well-used protein-protein interaction dataset for the inductive setting. Following~\cite{SuperGAT}, we repeated experiments $100$ times on Cora, CiteSeer, and PubMed with random seeds, and $30$ times on PPI, to produce reliable results.

We followed a general GAT model setting defined in the PyTorch Geometric~\cite{PYG} framework. We used a stacked two-layer \ours, with a learning-rate of $0.005$, optimized with an Adam optimizer~\cite{Adam}. We used the standard dot-product attention implementation when encoding one-hop structural information, which performed better than the linear attention proposed by GAT in our implementation. 

\paragraph{Baselines}
We selected the attention-base methods GAT~\cite{GAT}, CGAT~\cite{CGAT}, and SuperGAT$_\text{SD}$~\cite{SuperGAT} as baseline methods. Particularly, SuperGAT$_\text{SD}$ also uses the standard dot-product attention rather than that used in original GAT.  In addition, other GNN methods GCN~\cite{GCNs}, GraphSage~\cite{graphsage} and GCN+NS~\cite{GCN+NS} were also added for comparison.

\begin{table}[t]
\small
\centering
\caption{Node classification results on four benchmarks (accuracy for Cora, CiteSeer and PubMed; F1-score for PPI).}
\label{tab:main_node}
\resizebox{.98\linewidth}{!}{\scriptsize
\begin{tabular}{lllll}
\toprule
Model                & Cora  & CiteSeer & PubMed & PPI \\ \midrule
GCN  & 81.5   & 70.3 & 79.0 & 61.5$\pm$0.4 \\
GraphSage  & 82.1$\pm$0.6 & 71.9$\pm$0.9 & 78.0$\pm$0.7   & 59.0$\pm$1.2 \\ 
GCN+NS  & 83.7$\pm$1.4 & \textbf{74.1}$\pm$1.4 & -   & - \\
\midrule
GAT  & 83.0$\pm$0.7 & 72.5$\pm$0.7 & 79.0$\pm$0.4 & 72.2$\pm$0.6 \\
CGAT & 81.4$\pm$1.1 & 70.1$\pm$0.9 & 78.1$\pm$1.0 & 68.3$\pm$1.7 \\
SuperGAT$_\text{SD}$ & 82.7$\pm$0.6 & 72.5$\pm$0.8 & 81.3$\pm$0.5 & \textbf{74.4}$\pm$0.4 \\
\midrule
\ours & \textbf{84.6}$\pm$0.4 & 72.8$\pm$0.5 & \textbf{81.8}$\pm$0.3 & 73.7$\pm$0.3\\
\bottomrule
\end{tabular}}
\end{table}

\paragraph{Results}
The results are shown in Table~\ref{tab:main_node}, from which we can observe that \ours outperformed the attention-based methods on most datasets except PPI. Although SuperGAT$_\text{SD}$ had better performance on this dataset, we argue that it is no contradiction to incorporate SuperGAT$_\text{SD}$ as structural encoder into \ours to obtain a stronger model.  

Interestingly, the attention-based methods unanimously performed worse than the GCN-based method GCN+NS on CiteSeer. SuperGAT$_\text{SD}$ and CGAT even had the same or worse results compared with the original GAT. Nevertheless, we observe significant improvement from GAT to \ours. This result empirically demonstrates the strength of leveraging semantic neighbors.

\begin{table}[t]
% Numbers in \textbf{bold} represent the best results.
	\centering
	\caption{KG completion (entity prediction) results on FB15K-237 and WN18RR.}
	\resizebox{.98\linewidth}{!}{\scriptsize
	\begin{tabular}{lrrrrrr}
		\toprule
		\multirow{2}{*}{Model}  &  \multicolumn{3}{c}{FB15K-237} & \multicolumn{3}{c}{WN18RR} \\ \cmidrule(lr){2-4} \cmidrule(lr){5-7}
		& MRR & MR & Hits@1 & MRR & MR & Hits@1 \\ \midrule
% 		RESCAL   & .356 & 192 & .266 & .467 & 5408 & .439 \\
		TransE   & .310 & 199 & .218 & .232 & 1,706 & .061 \\
% 		DistMult & .342 & 177 & .249 & .451 & 4390 & .414 \\
% 		ComplEx  & .343 & 176 & .250 & .479 & 3817 & .441 \\
		RotatE   & .338 & 177 & .241 & .476 & 3,340 & .428 \\
		TuckER  & .358 & - & .266 &  .470 & - & .443 \\
        \midrule
% 		ConvE    & .338 & 176 & .247 & .439 & 4789 & .409 \\
		CoKE  & .364 & - & .272 &  .484 & - & .450 \\
		CompGCN  & .355 & 197 & .264 &  .479 & 3,533 & .443 \\
% 		RotH    & .344 & -   & .246 &  .496 & -    & .449 \\
		HittER     & .373 & 158 & .279 & .503 & 2,268 & .462 \\
		\midrule
		\ours & \textbf{.376} & \textbf{150} & \textbf{.281} & \textbf{.507} & \textbf{2,255} & \textbf{.465} \\
		\bottomrule
	\end{tabular}}
	\label{tab:main_kg}
\end{table}

\begin{table*}[t]
\small
\centering
\caption{Ablation study results on different datasets (metrics: MAE for ZINC; Accuracy for Cora, CiteSeer and PubMed; MR for FB15K-237 and WN18RR. $\uparrow$: higher is better; $\downarrow$: lower is better. $\times$: unavailable entry).}
\label{tab:ablation_study}
\resizebox{.98\linewidth}{!}{\scriptsize
\begin{tabular}{ccc|rrrrrr}
\toprule
Structural encoder & Semantic encoder & Semantic neighbor fetching & ZINC$\downarrow$ & Cora$\uparrow$ & CiteSeer$\uparrow$ & PubMed$\uparrow$ & FB15K-237$\downarrow$ & WN18RR$\downarrow$ \\ \midrule

$\surd$ & $\surd$ & $\surd$ & $\times$ & 84.6 & 72.8 & 81.8 & 150 & 2,255 \\
$\surd$ & $\surd$ &  & 0.113 & 82.6 & 72.7 & 78.1 & 151 & 2,305\\  
$\surd$ &   &   & 0.122 & 84.0 & 71.6 & 80.9 & 158 & 2,268\\
  & $\surd$ & $\surd$ & $\times$ & 84.1 & 72.5 & 81.5 & $\times$ & $\times$  \\
  & $\surd$ &  & 0.515 & 83.1 & 72.6 & 77.7 & $\times$ & $\times$\\

\bottomrule
\end{tabular}}
\end{table*}

\begin{figure*}[t]
	\centering
	\includegraphics[width=.95\linewidth]{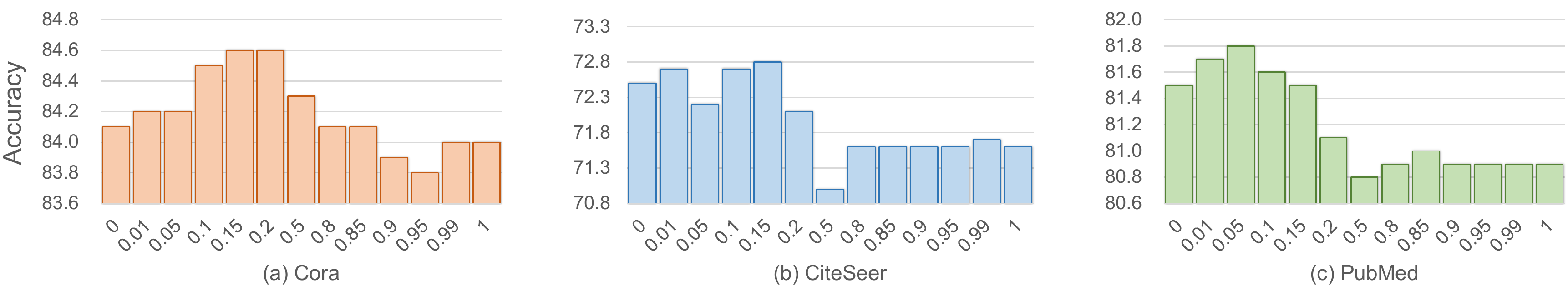}
	\caption{Accuracy on three citation network datasets, in term of hyper-parameter $\tau$ (average results of $7$ runs).}
	\label{fig:tau}
\end{figure*}

\subsection{KG Completion}
\paragraph{Settings} We conducted experiments on the entity prediction task to evaluate \ours for KG representation learning. The main target of entity prediction is to predict the subject entity (or object entity) for a given triple. We evaluated \ours on the current benchmarks FB15K-237~\cite{Node+LinkFeat} and WN18RR~\cite{ConvE}, which are sampled from the real-world KG Freebase~\cite{FreeBase} and WordNet~\cite{WordNet}, respectively.

The training setting mostly followed~\cite{HittER}. We used Adamax~\cite{Adam} optimizer with a learning rate of $0.01$. For each entity, we sampled at most $16$ semantic neighbors and updated them every $10$ epochs (rather than $1$ epoch) for efficiency. 

\paragraph{Baselines}
We chose the best-preforming entity prediction methods as our baselines: the TransE-family methods TransE~\cite{TransE}, RotatE~\cite{RotatE}, and TuckER~\cite{TuckER}, and the attention-based methods Coke~\cite{CoKE}, CompGCN~\cite{CompGCN}, and HittER~\cite{HittER}. Specifically, CoKE and HittER also leverage Transformer to encode structural information.

\paragraph{Results}
We report the main results on Table~\ref{tab:main_kg}. It can be clearly observed that \ours surpassed all the baselines across all datasets and metrics. The improvement on MR (mean rank) is most significant, which implies that \ours learned better embeddings for all entities, not only for top ones that valued by Hits@1. 
By comparing the results with those in Table~\ref{tab:main_graph_zinc} and Table~\ref{tab:main_node}, we actually find that the performance superiority of \ours is slim in the entity prediction task. We believe this is because KGs perhaps have the most complicated graph structures in real world. They have much more different edge types than networks or molecules. Encoding the structural information is thus more important for KGs. Even though, we still observe considerable improvement on two datasets.

\begin{figure*}[t]
	\centering
	\includegraphics[width=.85\linewidth]{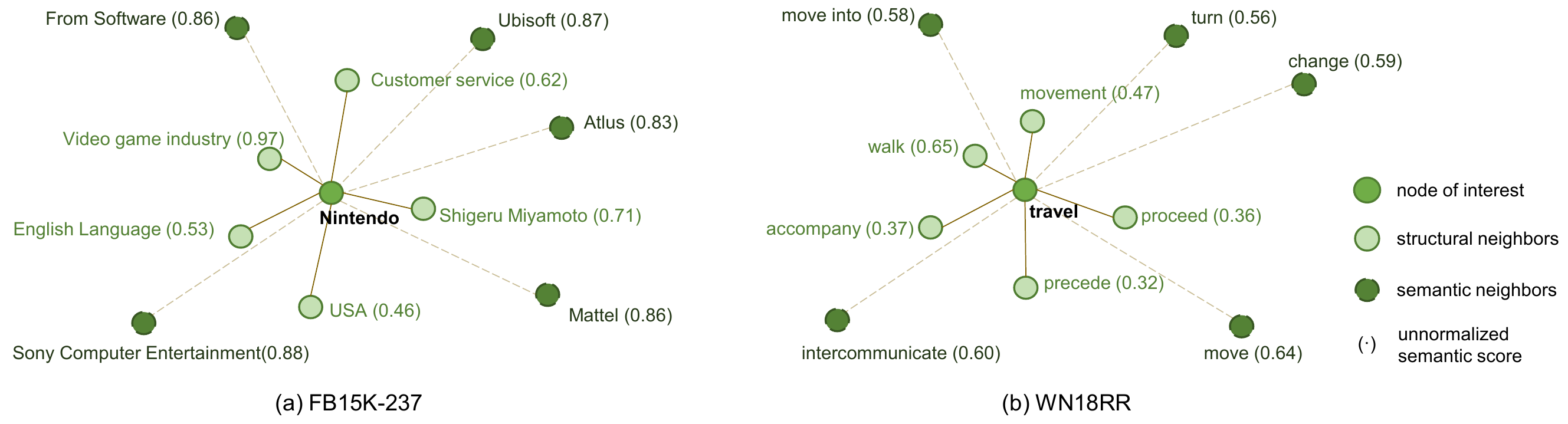}
	\caption{Examples of the semantic attention scores to the structural neighbors and semantic neighbors, respectively.}
	\label{fig:example}
\end{figure*}

\section{Further Analysis}
The superior performance inspires us to explore and evaluate \ours in depth. In this section, we design three experiments to help us better understand the idea of \ours.

\subsection{Is Every Module in \ours Useful?}
We first conducted ablation study experiments to verify the effectiveness of each module in \ours. We picked up six datasets in different representation learning tasks and present the results in Table~\ref{tab:ablation_study}. We removed the modules from \ours step-by-step while kept identical hyper-parameter setting along the experiments.

\paragraph{Semantic Neighbor Fetching} The semantic neighbor fetching loss is undoubtedly important to \ours. No matter combining two encoders or only using semantic encoder, integrating with the semantic fetching module had better performance in most cases. The improvement was most notable on PubMed, where it yielded $3.7\%$ and $3.8\%$ of accuracy increases, respectively. The mean rank results on WN18RR also got worse without the fetching loss.

\paragraph{Semantic Encoder} If we do not consider the semantic neighbor fetching loss, is semantic encoder itself useful for \ours? Unfortunately, we find the answer ambiguous. For Cora, PubMed, and WN18RR, when we did not employ the fetching loss, \ours with the semantic encoder performed much worse than \ours without the semantic encoder. But we observe that the situation reversed on CiteSeer and FB15K-237. In fact, when we only consider one-hop neighbors, the semantic encoder without fetching loss is just a ``minus'' attention layer. It may have its pros and cons compared with the standard dot-product attention layer on different datasets. In this sense, the semantic fetching loss is who endows the semantic encoder with the characteristic.

On the ZINC dataset, when it is capable of applying attention operations on the whole graph, the semantic encoder could estimate the semantic similarity between distant nodes without the help of the fetching loss. Therefore, we can see that the dual-encoding version of \ours greatly outperformed the structural encoder only version on ZINC. 
Overall, the effectiveness of the semantic encoder is conditioned: it must get in touch with the distant nodes.

\paragraph{Structural Encoder} The structural encoder also has merits. From the results of the $3$-rd and $5$-th rows in Table~\ref{tab:ablation_study}, we can find that it had better performance than the semantic encoder on most datasets except CiteSeer. The dot-product attention is worthy of analysis. We also noticed that the worst MAE on ZINC was obtained by only using the semantic encoder, due to the absence of all structural information.

\subsection{Does the Semantic Encoder Perform Better in Citation Networks?}
In Section~\ref{sec:methodology}, we mention that the semantic encoding may be more helpful in citation networks, where the semantically close neighbors are the key components to predict the label of the node of interest. In this section, we conducted experiment to verify this assumption empirically.
In \ours, we set a hyper-parameter $\tau$ to control the combination of the structural encoder and the semantic encoder, which can be written in the following equation:
\begin{equation}
    \rvh = \tau \rvh^\text{st} + (1-\tau) \rvh^\text{se},
\end{equation}
where $\rvh$, $\rvh^\text{st}$, $\rvh^\text{se}$ denote the combined output, structural output, and semantic output, respectively. By assigning different $\tau$, we can control the importance of each encoder in the combination.

The experimental results are shown in Figure~\ref{fig:tau}. Although the performance gap among three datasets was significant, we obverse a similar trend: the accuracy first increased steadily from $\tau=0$ to $\tau=0.1$, and then peaked around $\tau=0.15$, after which dropped rapidly until $\tau=1$.
When $\tau=0$ or $\tau=1$, we used only the semantic encoder or the structural encoder in training. The performance at $\tau=0$ is significantly better than that at $\tau=1$, which empirically demonstrates the advantage of leveraging semantic neighbor information. We further find that best performance was achieved at $\tau=0.15$, $0.15$, and $0.05$, respectively. This observation suggests that properly combining the output of two encoders may be the best choice. However, we should notice that, averaging the results ( i.e., $\tau=0.5$) was not a good idea on all dataset, as the semantic encoder always performed better than the structural encoder.

\subsection{How does the Semantic Encoder Help the Structural Encoder?}
It is worth exploring how the semantic encoder affects the structural encoder. In  Figure~\ref{fig:example}, we illustrate two examples on FB15K-237 and WN18RR, respectively. 
For the structural one-hop neighbors, we find that the scores calculated by the semantic encoder are in line with human intuition. For example, in the left figure, the entity \textit{USA} has a very low score although it is directly connected to \textit{Nintendo} by relation \textit{service\_location}. The verb \textit{precede} and \textit{accompany} also obtain relatively low scores in the right figure. These neighbors are not very related to the entities of interest from human perspective. On the other hand, some one-hop neighbors get high semantic scores, e.g., the well-known director \textit{Shigeru Miyamo} of \textit{Nintendo} in FB15K-237 and the verb \textit{walk} in WN18. They are the entities that should gain more attentions.
On the other hand, for the semantic neighbors, we can see that the exploited distant neighbors are closely related to the entity of interest. For example, \textit{Atlus} is an important game developer to \textit{Nintendo}. Aggregating such information may be helpful when the model is asked to predict the games related to \textit{Nintendo}. For the verb \textit{travel} in WN18RR, \textit{move} also shares many key features with it.
We also conducted experiment to analysis the effects of the semantic encoder to the structural encoder during the training phase. Limited by paper length, we refer the interested readers to Appendix~\ref{app:expr}.

\section{Conclusion and Future Work}
\label{sec:conclusion}
In this paper, we proposed a new Transformer architecture, called \ours, to deal with different types of graphs. In \ours, the structural encoder aggregates information from connected nodes while the semantic encoder seeks the distant nodes with useful semantics. The experimental results demonstrate the strong performance of \ours on three prevalent GNN tasks across $8$ benchmarks. We hope \ours can bring more insights and inspirations in developing unified Transformer architectures. In future, we plan to adapt \ours to NLP and CV areas.
\bibliography{reference_with_pagenumber}
\bibliographystyle{icml2022}

%%%%%%%%%%%%%%%%%%%%%%%%%%%%%%%%%%%%%%%%%%%%%%%%%%%%%%%%%%%%%%%%%%%%%%%%%%%%%%%
%%%%%%%%%%%%%%%%%%%%%%%%%%%%%%%%%%%%%%%%%%%%%%%%%%%%%%%%%%%%%%%%%%%%%%%%%%%%%%%
% DELETE THIS PART. DO NOT PLACE CONTENT AFTER THE REFERENCES!
%%%%%%%%%%%%%%%%%%%%%%%%%%%%%%%%%%%%%%%%%%%%%%%%%%%%%%%%%%%%%%%%%%%%%%%%%%%%%%%
%%%%%%%%%%%%%%%%%%%%%%%%%%%%%%%%%%%%%%%%%%%%%%%%%%%%%%%%%%%%%%%%%%%%%%%%%%%%%%%
\newpage
\appendix
\section{Position Embedding}
\label{app:pos}
In graphs, there are many important features that are used to identify different nodes. Thanks to learnable position embedding, these discrete features now can be encoding into embeddings and then combined with the raw embedding of nodes. 

Specifically, for graph representation learning task, we use the same method proposed by~\cite{Graphormer} to encode degree centrality of an arbitrary node $v_i$ as:
\begin{equation}
    \rvc_i = f_c(\text{deg}(v_i)),
\end{equation}
where $\text{deg}(v_i)$ denotes the degree of the node $v_i$, and $f_c: \sR \rightarrow \sR^h$ is the mapping function that converts the node degree to a learnable embedding. We further consider encoding the distances to different neighbors for the node of interest $v_i$ by the following equation:
\begin{equation}
    \rvd_{v_i, v_j} = f_d(\text{spd}(v_i, v_j)),
\end{equation}
where $\text{spd}(v_i, v_j)$ denotes the shortest path distance from $v_i$ to $v_j$, and $f_d: \sR \rightarrow \sR^h$ is a similar function that converts the distance to a learnable embedding.

For KG representation learning task, we follow~\cite{HittER} to encoder the edge types into node embeddings, which is implemented by an additional atom transformer $\gM_A$ to encode triples. Specifically, for a given triple $(v_i, r_{ij}, v_j)$ for the node of interest $v_i$, where $r_{ij}$ denote the edge type (i.e., relationship) between $v_i$ and $v_j$. The edge type information can be encoded by the following equation:
\begin{equation}
    \rve_{ij} = \gM_A([\rvc_A, \rvv_i, \rvr_{ij}, \rvv_j]),
\end{equation}
where $[\rvc_A, \rvv_i, \rvr_{ij}, \rvv_j]$ is the input embedding sequence. $\rvc_A$ is the virtual node for the atom Transformer, whose output represents the edge encoding embedding.

\section{Dataset Details}
\label{app:stat}

\begin{table*}[t]
\small
\centering
\caption{Dataset Statistics.}
\label{tab:dataset}
\resizebox{.98\linewidth}{!}{\scriptsize
\begin{tabular}{lrrrrrrrr}
\toprule
              & PCQM4M-LSC & ZINC & Cora  & CiteSeer & PubMed & PPI & FB15K-237 & WN18RR\\ \midrule
\# Graphs & 3,803,453 & 12,000 & 1 & 1 & 1 & 24 & 1 & 1 \\
\# Nodes & 53,814,542 & 277,920 & 2,708 & 3,327 & 19,717 & 56,944 &  14,951 & 40,943\\
\# Edges & 55,399,880 & 597,960 & 5,429 & 4,732 & 44,338 & 818,716 & 310,116 & 93,003\\
\# Edge-type & - & - & - & - & - & - & 237 & 11 \\
\# Task &  Regression & Regression & Classification & Classification & Classification & Classification & Classification & Classification\\
\# Classes & - & - & 7 & 6 & 3 & 121 & 14,951 & 40,943 \\\bottomrule
\end{tabular}}
\end{table*}

We present the overall dataset statistics in Table~\ref{tab:dataset}.

\subsection{Graph Property Prediction}
For PCQM4M-LSC, the model is asked to predict the DFT (density functional theory)-calculated HOMO-LUMO energy gap of given molecules. It contains more than 3.8M 2D molecular graphs as input, which is especially appropriate to evaluate the performance of model in large scale scenarios. On other hand, ZINC is a relative small datasets, where the main target is to predict the graph property regression for constrained solubility. It is one of the most popular real-world molecular datasets for graph representation learning.

\subsection{Node Classfication}
Cora, CiteSeer and PubMed are three citation network datasets proposed by~\cite{Cora}. They are typically used for transductive node classification task. PPI on the other hand is used for inductive evaluation. It consists of $24$ graphs, with $20$ graphs for training, and $2$ for validation and $2$ for testing. All four datasets are the prevalent benchmarks used for node classification.

\subsection{KG Completion}
FB15K-237 is the revised version of the original FB15K dataset~\cite{TransE} that was used as entity prediction benchmark in last ten years. However, recent studies~\cite{Node+LinkFeat,ConvE} find that the original FB15K contains a large proportion of redundant data, some of which may incur testing data leakage, the same to another well-used dataset WN18. Therefore, most latest studies only use the revised datasets FB15K-237 and WN18RR for evaluation. FB15K-237 has more different relationships, while WN18RR is more sparse and has more different entities.

\section{Analysis on the Training Process}
\label{app:expr}

\begin{figure}[t]
	\centering
	\includegraphics[width=.8\linewidth]{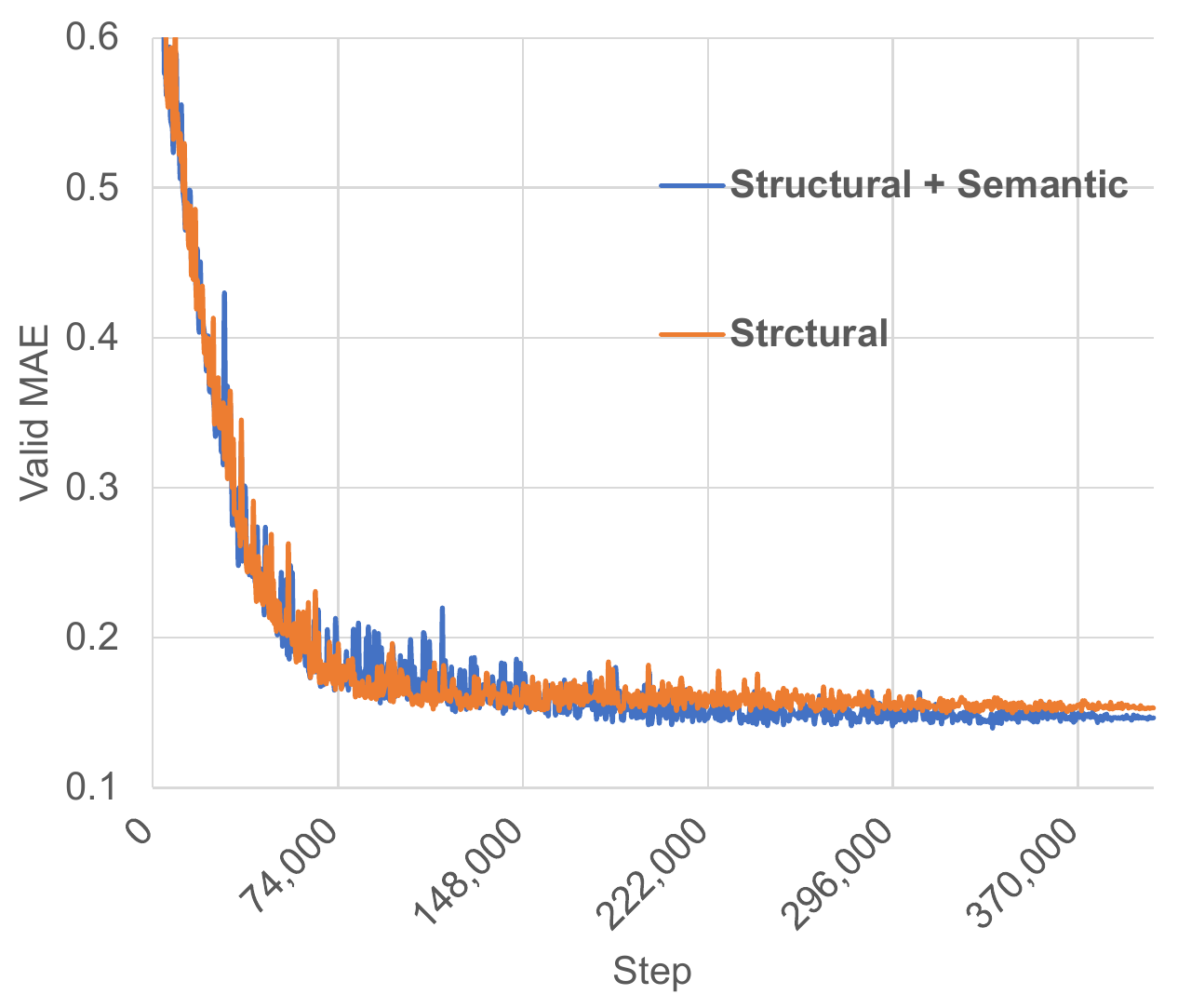}
	\caption{The validate MAE results on the ZINC dataset, in term of training step.}
	\label{fig:zinc}
\end{figure}

\begin{figure}[t]
	\centering
	\includegraphics[width=.8\linewidth]{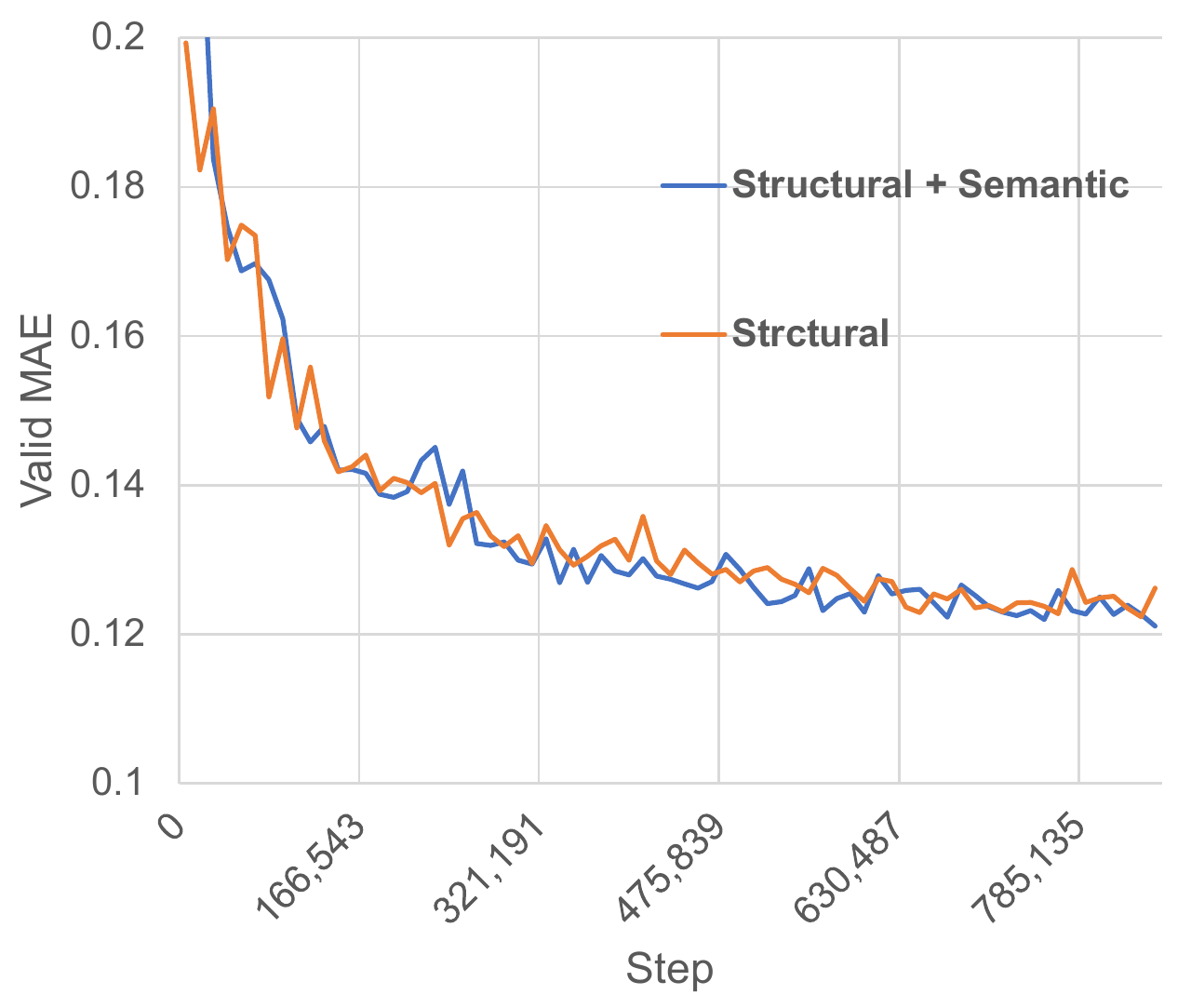}
	\caption{The validate MAE results on the PCQM4M-LSC dataset, in term of training step.}
	\label{fig:lsc}
\end{figure}

We conducted experiments to compare the performance of structural encoder with and without the semantic encoder, in term of training steps.

We depict the valid MAE results on ZINC in Figure~\ref{fig:zinc}. At the beginning of the training, we find that the two methods do not have a visible performance gap. The curves are tightly overlapped during step $0$ to $74,000$. 

As the performance starts to be converged, i.e., step $74,000$ to $148,000$, only using structural encoder is better than combining the two encoders. However, as the valid MAE tends to be stable, the dual-encoding \ours gradually outperforms the single structural encoder.

It is worth-noting that the turning point appears at the performance starts to be converged, where the input embeddings also approach the ideal positions. Therefore, the semantic similarity among embeddings can be estimated more precisely, and contribute to a better semantic encoder. Therefore, \ours can obtain a lower MAE than the single strctural encoder. We present the result on PCQM4M-LSC in Figure~\ref{fig:lsc}, which supports the same conclusion.
\end{document}